\documentclass{article}

  \usepackage[preprint, nonatbib]{neurips_2020}

\usepackage[utf8]{inputenc} 
\usepackage[T1]{fontenc}    
\usepackage{hyperref}       
\usepackage{url}            
\usepackage{booktabs}       
\usepackage{amsfonts}       
\usepackage{nicefrac}       
\usepackage{microtype}      
\usepackage{algorithm}
\usepackage{algorithmic}
\usepackage{amsthm}
\usepackage{amssymb}
\usepackage{amsmath}
\usepackage{pgfplots}
\usepgfplotslibrary{groupplots}
\usepgfplotslibrary{fillbetween}
\usepackage{subfig}
\usepackage{graphicx}

\pgfplotsset{every axis/.append style={
                    legend style={font=\tiny,line width=.5pt,mark size=.6pt},
                    }}

\newtheorem{theorem}{Theorem}
\newtheorem{definition}[theorem]{Definition}
\newtheorem{corollary}[theorem]{Corollary}

\newtheorem{Lemma}[theorem]{Lemma}

\newtheorem{Remark}[theorem]{Remark}

\newcommand{\asp}{A^{S^{'}}}
\newcommand{\as}{A^{S}}
\newcommand{\myparagraph}[1]{\noindent {\bf #1}}

\renewcommand{\vec}[1]{{\mathbf{#1}}}

\DeclareMathOperator*{\argmin}{arg\,min}
\DeclareMathOperator*{\Lap}{Lap}

\newcommand{\abs}[1]{\left|#1 \right|}

\renewcommand{\H}{\mathcal{H}}

\newcommand{\reals}{\mathbb{R}}
\let\eps\epsilon

\title{Stability Enhanced Privacy and  Applications in Private Stochastic Gradient Descent}

\author{%
 Lauren Watson \\
School of Informatics \\ University of Edinburgh\\
 \hspace{-1mm}\texttt{lauren.watson@ed.ac.uk} \\
  \And
Benedek Rozemberczki  \\
School of Informatics \\ University of Edinburgh\\
  \hspace{-2mm}\texttt{benedek.rozemberczki@ed.ac.uk} \\
  \And
    Rik Sarkar \\
School of Informatics \\ University of Edinburgh\\   
  \hspace{-1mm}\texttt{rsarkar@inf.ed.ac.uk} \\
}

\begin{document}

\maketitle

\begin{abstract}

Private machine learning involves addition of noise while training, resulting in lower accuracy. Intuitively, greater stability can imply greater privacy and  improve this privacy-utility tradeoff. We study this role of stability in private empirical risk minimization, where differential privacy is achieved by output perturbation, and establish a corresponding theoretical result showing that for strongly-convex loss functions, an algorithm with uniform stability of $\beta$ implies a bound of $O(\sqrt{\beta})$ on the scale of noise required for differential privacy. 

The result applies to both explicit regularization and to implicitly stabilized ERM, such as adaptations of Stochastic Gradient Descent that are known to be stable. Thus, it generalizes recent results that improve privacy through modifications to SGD, and establishes stability as the unifying perspective. It implies new privacy guarantees for optimizations with uniform stability guarantees, where a corresponding differential privacy guarantee was previously not known. Experimental results validate the utility of stability enhanced privacy in several problems, including application of elastic nets and feature selection. 

\end{abstract}


\section{Introduction}

Privacy is important in the widespread use of machine learning, as learning algorithms are increasingly applied to sensitive data. When sensitive information is present in the training data, the model output by the training process can reflect the presence of specific data items, and thus leak private information~\cite{song2017machine}.

Differential privacy~\cite{Dwork06differentialprivacy} has emerged as the gold standard definition of statistical privacy guarantees for machine learning. The corresponding mechanisms operate by adding random noise to the training  process. Greater noise ensures greater privacy, but comes at the cost of greater loss of accuracy. Differentially Private Empirical Risk Minimization has been a topic of extensive study in the last decade~\cite{chaudhuri2009privacy, chaudhuri2011differentially,kifer2012private,bassily2014private, Wangprivacy, wang2017}.

An algorithm is called stable~\cite{bousquet2002stability} when it is guaranteed to have only a small change in the loss, on a small change to the training dataset. Stability is known to be closely related to generalization properties~\cite{rogers1978finite,elisseeff2005stability,mukherjee2006learning}. It can be incorporated directly into learning objectives by using a regularization term, or, as has been shown in recent works, optimization algorithms such as stochastic gradient descent can be made more stable by various common modifications such as gradient clipping, dropout, batch normalization, smaller step sizes  etc~\cite{hardt2015,santurkar2018does}.

In this paper we show that uniform stability, defined as a bound on the change in the loss function, in fact implies a bound on the change in the actual model output by the learning function, under suitable convexity conditions. This result implies greater privacy for the same level of noise for any uniformly stable algorithm, and thus generalises recent results that use specific algorithmic modifications to attain differential privacy (e.g.~\cite{rubinstein2012learning,wu2017bolt}). The result implies guarantees of greater privacy for any modifications that improve the uniform stability of a learning algorithm. Our analysis applies to output perturbation approaches to privacy, and thus can be used without modification of existing implementations in an add-on manner suggested in~\cite{wu2017bolt}.

We describe a specific variant of dropout that increases uniform stability, thus improving privacy. Table~\ref{tab:stability} presents other versions of optimizations such as stochastic gradient descent with known uniform stability bounds. As the table shows, in many of these cases, corresponding privacy results have not been derived in previous works, and we can now guarantee improved stability-enhanced-privacy for these versions.

We discuss the trade-off between the empirical error, training error and error due to privacy noise induced by stability. As a practical application area, we discuss classification and feature selection in elastic net optimizations. Experimental results on multiple datasets show that using the relation between stability and privacy, same levels of privacy can be obtained at substantially higher accuracy.


\section{Preliminaries}
\vspace*{-2mm}
Empirical risk minimization refers to finding the model that minimizes training loss over a given dataset. Given a training set $S =\{(\mathbf{x}_i, y_i)\in \mathcal{X} \times \mathcal{Y} $ for $i=\{1, ..., n\}$\}, suppose $c:\mathcal{Y}\times\mathcal{Y}\rightarrow \reals^+$ is a cost  function. Suppose $\H$ is a hypothesis class, and each hypothesis $h:\mathcal{X}\rightarrow \mathcal{Y}$ can be represented by a vector $\vec{w}$ of paramteters. The loss of $h$ on an item $s=(x,y)$ is given by $\ell(h,s) = c(h(x),y)$.

The objective is to find the vector $\vec{w}_S$ describing $\as$ -- the hypothesis that minimizes the average loss over $S$:
\begin{equation}
 \as=\argmin_{h\in \H}\frac{1}{n}\sum_{i=1}^n \ell(h, s_i)
\end{equation}
Thus, finding $\vec{w}_S$ is equivalent to finding the best-on-average model for the training data. We use $A(S)$ and $\as$ interchangeably at times, in order to emphasize that the algorithm is a function of $S$.

A learning algorithm is stable if the loss of the output model or hypothesis described by $\vec{w}_S$ changes only slightly on a small change in input~\cite{bousquet2002stability}. We first define neighboring datasets as ones that differ from one another in at most one element.

\begin{definition}[\bf Neighbouring Databases]
Two databases $S, S'$ are neighbouring if $H(S, S') \leq 1$, where $H(\cdot, \cdot)$ represents the hamming distance.
\end{definition}

We are interested in {\em Uniform Stability}~\cite{bousquet2002stability}, which requires that between neighboring datasets, the change in loss is bounded by $\beta$. Note that greater stability corresponds to smaller values of $\beta$.

\begin{definition}[\bf $\beta$-Uniform Stability~\cite{bousquet2002stability}]
  An algorithm $A$ satisfies $\beta$-uniform stability with respect to the loss function $\ell$ if for neighboring datasets $S, S'\sim \mathcal{D}$ and  for every datapoint $s\in S$, \mbox{
$
    |\ell(A(S), s)-\ell(A(S'), s)|\leq \beta
$}
where $\mathcal{D}$ is the (possibly unknown) distribution from which the samples are drawn.
\end{definition}

Differential privacy is the guarantee that changing one element of a database does not change the output probabilities of a probabilistic mechanism by more than a constant factor:
\begin{definition}[\bf Differential Privacy~\cite{Dwork06differentialprivacy}]
  A randomized algorithm $M$ satisfies $\epsilon$-differential privacy if for all neighbouring databases $S$ and $S'$ and for all possible outputs $O\subseteq \text{Range}(M)$, \mbox{
  $
      \Pr[M(S) \in O] \leq e^{\epsilon} \cdot \Pr[M(S') \in O].
  $}
\end{definition}

Differential privacy is usually achieved by adding noise. The scale, or variance, of the noise depends on the function $f$ being computed. The {\em sensitivity} of a deterministic vector valued function $f$ is given by $\Delta f = \max|f(S) - f(S^\prime)|$, with the maximum taken over all possible neighboring $(S, S^\prime)$. Sensitivity uses the $L_1$ or the $L_2$ norm depending on the privacy mechanism~\cite{dwork2006calibrating, wu2017bolt}.

Differential privacy can be achieved by returning $f(S) + u$, where $u$ is sampled from distribution $P(u)\propto \exp\left(-\eps|u|/\Delta f\right)$. This is the Laplace distribution, sometimes written as $\Lap\left(\Delta f/\eps\right)$\footnote{$\Lap(b)$ is the Laplace distribution with mean $0$ and variance $2b^2$. $b$ is called the {\em scale of the distribution}.}. Several variants have been developed for differentially private noise addition to optimization algorithms such as stochastic gradient descent. See~\cite{wang2017differentially} for a comparison. The approach described above is called {\em Output perturbation} -- where SGD operates normally, and noise is added to the output. {\em Objective perturbation} was proposed by~\cite{chaudhuri2011differentially}, where the objective function itself is perturbed. In gradient perturbation~\cite{chaudhuri2011differentially}, the computed gradient at each step is perturbed by noise.

\subsection{Related Work}

\myparagraph{Uniform Stability.} Uniform stability was proposed by~\cite{bousquet2001} and further developed in~\cite{bousquet2002stability}.  
The study of algorithmic stability has been mainly used to bound the generalization error of learning algorithms~\cite{rogers1978finite}. 
Many results on uniform stability use stability of the expected loss of randomized algorithms~\cite{elisseeff2005stability}, providing average-case~\cite{hardt2015, mou2018sgld, singh2016swapout} but not necessarily the worst-case uniform stability bounds. 
Recently, \cite{feldman2018generalization} provided tighter generalization bounds for uniformly stable algorithms in this setting. Some modifications of learning algorithms such as stochastic gradient descent have been shown to induce uniform  stability~\cite{hardt2015, wu2017bolt}. Similar observations have been made for multi-task learning under mild assumptions~\cite{liu2016algorithm}.

\myparagraph{Private ERM.} A differentially private method for logistic regression was described in~\cite{chaudhuri2009privacy}.
This work was followed by~\cite{chaudhuri2011differentially}, which extended their results to the general setting of regularized ERM algorithms. Stability based analysis inspired from~\cite{bousquet2002stability} is also used in~\cite{rubinstein2012learning} for private soft-margin support vector machines (SVMs). 
Bounds for private ERM in more general settings are discussed by ~\cite{kifer2012private, bassily2014private, jain2014near} and~\cite{wang2017}. 

\myparagraph{Stability and Privacy.} It is generally known that differential privacy implies stability. For example, if an algorithm is $\epsilon$-differentially private then it is $2\epsilon$-uniform stable~\cite{Wangprivacy}. However, the other direction, of how improving the stability of an algorithm influences privacy, is less clear. In comparison to these works, we establish a relation between uniform stability and sensitivity, therefore obtaining differentially private algorithms in terms of uniform stability. As a result, it can be applied to any algorithms that applies noise based on sensitivity. In particular, this provides an approach for reducing the sensitivity factor in privacy profiles as studied in amplification by subsampling~\cite{balle2018privacy}.
Our results also demonstrate the relationship between uniform stability and \emph{uniform argument stability}~\cite{liu2017algorithmic} for strongly convex loss functions.


\section{Privacy via Stability}
\vspace*{-2mm}

We now demonstrate how sensitivity can be bounded using uniform stability. Let us first examine the case of regularized empirical risk minimization, where $A^S$ minimizes the loss function plus a term penalizing large weights:
\begin{equation}
    \frac{1}{n}\sum_{i=1}^n \ell(A^S(\mathbf{x}_i), y_i) + \frac{\lambda}{2} ||\textbf{w}_S||_2^2
\end{equation}
Regularization prevents the model from overfitting to the training data, therefore improving the generalization ability of the model. The relationship between sensitivity of the output model and stability follows in this case by a direct extension of existing results on the sensitivity and stability of regularized ERM~\cite{chaudhuri2011differentially, rubinstein2012learning, bousquet2002stability}.

\begin{theorem}
Let $A$ denote a regularized empirical risk minimization algorithm, over dataset $S$, with convex loss function $\ell(\cdot, \cdot)$ which has Lipschitz constant $L$. Let $\mathcal{H}$ be a RKHS with a $d$-dimensional feature mapping with bounded norm $k(\mathbf{x}, \mathbf{x})\leq \kappa^2$ where $\mathbf{x}\in \mathbb{R}^d$. Then for all neighbouring databases $S, S'$, the sensitivity of the weights output by $A$ is bounded as $||\textbf{w}_{S}-\textbf{w}_{S'}||_2\leq \frac{4L\kappa}{n\lambda}$.
\label{thm:ruben}
\end{theorem}
The proof of Theorem~\ref{thm:ruben} follows similar arguments as~\cite{rubinstein2012learning, bousquet2002stability} and directly implies that increasing the regularization in an empirical risk minimization algorithm decreases the output weight sensitivity of that algorithm. Note that the corresponding $L_1$-sensitivity is then bounded by $\frac{4L\kappa\sqrt{d}}{n\lambda}$.

$L_2$-regularized ERM algorithms are known to be uniformly stable with $\beta = \frac{2L^2\kappa^2}{n\lambda}$~\cite{scholkopf2001learning, bousquet2002stability}, directly implying the following relationship between stability and sensitivity in this context:
\begin{corollary}
Let $A$ denote a regularized empirical risk minimization algorithm with regularization parameter $\lambda$, over dataset $S$, with convex loss function $\ell$ which has Lipschitz constant $L$ and satisfies $\beta$-uniform stability. Let $\mathcal{H}$ be a RKHS with a $d$-dimensional feature mapping with bounded norm $k(\mathbf{x}, \mathbf{x})\leq \kappa^2$ where $\mathbf{x}\in \mathbb{R}^d$. Then the  sensitivity of A is bounded by:$||\textbf{w}_{S}-\textbf{w}_{S'}||_2\leq \frac{2\beta}{L\kappa}$

\end{corollary}

We now explore the more general case of empirical risk minimization for strongly-convex loss functions, which is a common setting in work on private empirical risk minimization~\cite{chaudhuri2011differentially, kifer2012private,bassily2014private}.

\begin{definition}{\textbf{$\lambda$-Strong Convexity:}}
  $\ell$ satisfies $\lambda$-strong convexity if for all $s\in S$:
\[
    \ell(A(S), s)\geq \ell(A(S'), s) + \nabla \ell(A(S'), s)^T(\mathbf{w}_S-\mathbf{w}_{S'}) + \frac{\lambda}{2}||\mathbf{w}_S-\mathbf{w}_{S'}||_2^2
\]\vspace*{-3mm}
\end{definition}

Strong convexity implies that the second derivative of a function is at least a positive constant, and that the growth of the function is lower bounded by a quadratic.
\begin{theorem}
Let $A$ denote a $\beta$-uniformly stable empirical risk minimization algorithm over dataset $S$, with $\lambda$-strongly convex loss function $\ell$ . The output sensitivity of $A$ is bounded by:
\[
||\textbf{w}_S-\textbf{w}_{S'}||_2 \leq \sqrt{\frac{2\beta}{\lambda}}.
\]\vspace*{-3mm}

\label{thm:stability}
\end{theorem}

The corresponding $L_1$-sensitivity is then bounded by $\sqrt{\frac{2d\beta}{\lambda}}$. 
The theorem implies that the sensitivity of the weights output by an algorithm $A$ can be bounded using uniform stability. In other words, the sensitivity of the \emph{weights} learned via an algorithm $A$ can be bounded using the sensitivity of its \emph{loss}.

Note that a convex loss function with $L_2$-regularization parameter $\lambda$ is $\lambda$-strongly convex. However, the theorem applies more generally where strong convexity is implicit, and not due to an added regularization term.

Theorem~\ref{thm:stability} implies off-the-shelf sensitivity bounds for uniformly stable algorithms, such as those listed in Table~\ref{tab:stability}. In particular, it provides private counterparts for variants of stochastic gradient descent, such as Nesterov accelerated gradient descent~\cite{dozat2016incorporating}. The natural way to utilize this result is through the use of output perturbation (See Algorithm~\ref{alg:output}).
\begin{algorithm}
\caption{Output Perturbation via stability}
\begin{algorithmic}[1]
  \scriptsize
  \STATE Input: $S=\{(\mathbf{x}_i, y_i)\}$ for $i\in[1:n]$, loss function $\ell$, sensitivity of output $\Delta w=\sqrt{\frac{2\beta}{\lambda}}$ and privacy level $\epsilon$.
  \STATE  $\mathbf{w}_S \leftarrow $ minimizer of $\ell(A, S)$.
  \STATE Return: $\mathbf{w}_S+\nu$ where $\nu\sim Lap\left(\frac{\Delta w}{\epsilon}\right)$.
  \label{alg:output}
\end{algorithmic}
\end{algorithm}\vspace*{-2mm}

\begin{Remark}\label{rem:conv}
For iterative methods, the true minimum loss may not be reached in a finite number of steps. 
Suppose the weights output after $T$ steps is  $\widetilde{\mathbf{w}}_S$. Let the weight convergence rate be given by $||\widetilde{\mathbf{w}}_S-\mathbf{w}_S||_2\leq \delta_{conv}$. The sensitivity of this method is then bounded as follows:
$||\widetilde{\mathbf{w}}_S-\widetilde{\mathbf{w}}_{S'}||_2
    \leq 2\delta_{conv}+||\mathbf{w}_{S}-\mathbf{w}_{S'}||_2.$
 Alternatively, let the loss convergence rate be given by $|\widetilde{\ell}(A^S, s)-\ell(A^S, s)|\leq \gamma_{conv}$. In a similar manner to Theorem~\ref{thm:stability}, for any strongly-convex loss function, the weight convergence rate $\delta_{conv}$ can be replaced with the loss convergence rate by observing that $\delta_{conv}\leq \sqrt{2\gamma_{conv}/\lambda}$.

\end{Remark}

\section{The Stability-Privacy Trade-off}
Bounding sensitivity via stability sheds light on the trade-off between empirical accuracy for the underlying model and the noise required to provide privacy.
We first describe empirical, generalization and privacy error in terms of stability. We then provide examples of parameters which can be used to tune stability for algorithms such as stochastic gradient descent and note their privacy enhancing effect and demonstrate how a variant of dropout can be used to improve stability.
\subsection{Characterizing Error via Stability}\label{sec:tradeoff}

Stability describes the trade-off between achieving low training error and low generalization error in non-private ERM~\cite{bousquet2002stability}. In an output perturbation scenario, where noise is added proportional to the level of sensitivity determined by the uniform stability of the algorithm, stability also controls excess error due to the noise added for privacy.

Denote the empirical loss for an empirical risk minimization algorithm $A$ by $L(A, S)=\frac{1}{n}\sum_{i=1}^n \ell(A^S,s_i)$, the expected population loss by  $\hat{L}_{\mathcal{D}}(A)=\mathbb{E}_{s\sim D}[\ell(A, s))]$ and the loss achieved by $A$ with output perturbation by $L_{priv}(A,S)$.

\begin{definition}[\bf Excess generalization error]
The expected excess generalization error of an algorithm $A$ trained over database $S\sim\mathcal{D}$ is defined as $\delta_{gen}=\mathbb{E}_{S\sim \mathcal{D}}[\hat{L}_{\mathcal{D}}(A)-L(A,S)]$.
\end{definition}
Excess generalization loss decreases as stability increases, with $|\delta_{gen}|\leq \beta$~\cite{bousquet2002stability}.
\begin{definition}[\bf Excess privacy error]
Excess privacy error (privacy loss) is defined as $
    \delta_{priv}=\mathbb{E}_S[|L_{priv}(A,S)-L(A,S)|]$.
\end{definition}
\begin{Lemma}\label{lemma:priverror}
Assume $A$ is a $\beta$-uniformly stable learning algorithm, with $L$-lipschitz loss $\ell$ and a $d$-dimensional input space. Denote the $\epsilon$-differentially private counterpart of this algorithm, obtained by applying output perturbation with sensitivity given by Theorem~\ref{thm:stability}, by $A_{priv}$. The excess privacy error introduced by $A_{priv}$ is given by $\delta{priv}\leq \frac{Ld}{\epsilon }\sqrt{\frac{2\beta}{\lambda}}$
\end{Lemma}
Lemma~\ref{lemma:priverror} demonstrates that privacy error decreases as stability increases. In contrast, empirical error often increases as stability increases, as for highly stable algorithms closeness of fit to the training data is reduced. For example, denote the minimum empirical loss achieved by an iterative algorithm in a fixed number of steps as $\tilde{L}(A,S)$. Given a $\beta$-uniformly stable and $\sigma$-smooth iterative algorithm $A$ with convex loss function $\ell$, then  $\delta_{emp}\geq O(\frac{\sigma}{n\beta})$~\cite{chen2018stability}.
\begin{definition}[\bf Excess empirical error]
The expected empirical (excess training) error of an algorithm $A$ trained over database $S$ is defined as
$\delta_{emp}=\mathbb{E}_S[\tilde{L}(A,S)-L(A,S)]$.
\end{definition}

In this case, increasing stability reduces bounds for both generalization and privacy error, but increases empirical error. For small $\epsilon$, modest gains in empirical error afforded by a weaker level of stability (e.g. less regularized) could be outweighed by the corresponding increase in privacy error. This suggests that output perturbed private learning algorithms may require a higher level of regularization than their non-private counterparts. This claim is empirically supported by both the results presented in Section~\ref{sec:exps} and empirical results presented in seminal work on private ERM~\cite{chaudhuri2011differentially}. Section~\ref{sc:examples} provides examples of parameters which can be used to tune uniform stability for various algorithms.

\subsection{Stability Enhancing Methods}\label{sc:examples}

Table~\ref{tab:stability} provides examples of uniformly stable algorithms, alongside parameters which can be used to control their stability. See Appendix~\ref{sec:averagestability} for average-case uniformly stable algorithms. Typically, the $L_2$-regularization parameter $\lambda$ is used to tune the trade-off between underlying model fit and sensitivity. However, as demonstrated by Table~\ref{tab:stability}, various other parameters can be used to control stability (and thus sensitivity) in different scenarios. Many of these methods can be viewed as performing implicit regularization and are widely used across machine learning. As the table shows, in many versions of optimization, there are known stability bounds, but no corresponding privacy bound. In these cases, we can now claim improved privacy.

\begin{table}
\begin{tabular}{l||c|c|c|c}
Method & Parameters & Stability &  \begin{tabular}[c]{@{}l@{}} Prev. Work  \\ (Output Pert.) \end{tabular} & \begin{tabular}[c]{@{}l@{}} Prev. Work  \\ (Obj. Pert.) \end{tabular}\\ \hline \hline
Regularized ERM~\cite{bousquet2001} & $n, \lambda, L$ & $O(\frac{L^2}{n\lambda})$ & \cite{chaudhuri2011differentially, rubinstein2012learning}&\cite{chaudhuri2011differentially, rubinstein2012learning}\\
\hline
    SGD - Steps~\cite{hardt2015}     &  $T$ Steps & $O(T)$          & \cite{wu2017bolt, zhang2020privacy}   &   \cite{bassily2019private}         \\
     SGD - Step Size~\cite{hardt2015}     &  Step Size $\alpha$ & $O(\alpha)$          & \cite{wu2017bolt,zhang2020privacy}           & -      \\
     SGD - Model Averaging~\cite{hardt2015}  &$\alpha_t$ & $O(\sum_{t=1}^T\alpha_t)$ &\cite{wu2017bolt}&-\\
    SGD - Minibatch Training & Batch size $b$& $O(\frac{1}{b})$ &\cite{wu2017bolt}&- \\
     SGD - s-Dropout(Sec.~\ref{sec:drop})    & Rate $s$ &   $O(s)$      & -  &-               \\

    SGD - Grad. Clipping~\cite{hardt2015} & Grad. $G$& $O(min(G, L))$ &-&\cite{abadi2016deep} \\
    SGD - Batch Norm.\footnotemark~\cite{santurkar2018does} & $\gamma,\sigma$& $O(\frac{\gamma^2}{\sigma^2})$& - & - \\
    \hline
    Nesterov Acc. GD~\cite{chen2018stability} & $T$ Steps & $O(T^2)$ & - & -\\
    Heavy Ball Method~\cite{chen2018stability} &  $\gamma$&$O(\frac{1}{1-\sqrt{\gamma}})$&- & -\\

\hline
    Multi-Task Learning\footnotemark~\cite{liu2016algorithm} & $T$ Tasks & $O(\frac{1}{T})$ & - & \cite{zhang2020privacy}\\
    Elastic Net~\cite{xu2011sparse} & $ \lambda, \gamma$ & $O(\frac{1}{\lambda\gamma})$ & - & \cite{poggio2009sufficient,yu2014differentially}\\
    Bridge Regression~\cite{poggio2009sufficient} & $p$ & $O(\frac{1}{p(p-1)}\frac{1}{\lambda}^{\frac{2-p}{p}})$ & - & - \\
   $k$-partite ranking~\cite{gao2013stability}    &$\lambda$&  $O(\frac{1}{\lambda})$ &- &-\\
    \hline

\end{tabular}
\vspace{1mm}
\caption{Uniformly stable algorithms and the parameters that control their stability. Related work included in the last two columns indicate if the relationship to the described parameter has been previously used in the context of output perturbation or objective perturbation. Theorem~\ref{thm:stability} implies new privacy bounds in cases where this relationship was not known previously.}
\label{tab:stability}
\vspace*{-6mm}
\end{table}
\addtocounter{footnote}{-1}
\footnotetext{Batch normalization has been shown to improve the Lipschitzness of the loss function~\cite{santurkar2018does}. This improvement can be substituted into existing worst-case stability results for SGD~\cite{wu2017bolt}. }
\addtocounter{footnote}{1}
 \footnotetext{The uniform stability result of~\cite{liu2016algorithm} can be applied to a single task used in the final output of the MTL model where the other tasks regularizes the model. \cite{gupta2016differentially} used output perturbation for MTLR. }

Corollary~\ref{cor:enhance} implies that identifying stability enhancing actions within a private machine learning algorithm, such as those listed in Table~\ref{tab:stability}, amplifies its privacy guarantee without requiring more noise. If $\beta$ is the best known stability guarantee for algorithm $A$, then a Laplace noise of scale $\frac{O(\sqrt{\beta})}{\eps}$ guarantees   $\eps$-differential privacy. If the stability guarantee of $A$ is improved, for example, due to better analysis, then the privacy guarantee improves correspondingly without additional noise:

\begin{corollary}[\bf Privacy enhancement] Suppose $A$ has known guarantee of $\beta_1$-uniform stability, and Laplace perturbation of scale $b$ guarantees $\eps$-differential privacy. If stability guarantee of $A$ improves to $\beta_2 < \beta_1$, then under the same perturbation strategy, the privacy guarantee improves to $\sqrt{\frac{\beta_2}{\beta_1}}\cdot \eps$-differential privacy.

\label{cor:enhance}
\end{corollary}

\myparagraph{Increasing Stability via Dropout.}\label{sec:drop}
We now provide an example of a variant of dropout which can be used to increase stability. In stochastic gradient descent, dropout~\cite{srivastava14a} is equivalent to updating the weights using $D\nabla \ell$ as opposed to $\nabla \ell$, where $D$ is a randomized `mask' setting some values of $\nabla \ell$ to 0~\cite{hardt2015}.

We present a modified definition of dropout as follows:

\begin{definition}[\bf s-Dropout]\label{def:sdrop}
An \emph{s-dropout} operator is a randomized map $D: \Omega \rightarrow \Omega$ with dropout
rate $s\leq 1$ such that for every $v \in \Omega$, $||D(v)||\leq s||v||$\end{definition}

The difference here is that the contraction to the vector is by a factor of $s$ or less in all cases, rather than in expectation. This operator can be implemented for any given constant $s$ by randomly dropping components of $v$ until the norm is smaller than $s||v||$.

The Lipschitz constant of the gradient update is then reduced from $L$ to $sL$~\cite{hardt2015} and the gradient update with step-size $\eta$ is then $\eta s L$-bounded, as opposed to $\eta L$-bounded. As the stability of SGD is $O(L)$~\cite{wu2017bolt}, the stability is improved by a factor of $s$.

\section{Applications of Privacy via Stability}

In this section, we apply privacy via stability to the problem of private classification and feature selection, providing an output perturbation approach to private elastic-net regularized algorithms~\cite{zou2005regularization}.

\myparagraph{Private Classification with Elastic-Net}
An alternative to L2-regularization is to penalize the L1-norm of the obtained weights. This alternative (called LASSO) encourages sparsity and implicit feature selection, with many learned weights equal to 0~\cite{tibshirani1996regression}. Despite the  advantages of sparsity, L1-regularized models under-perform L2-regularized models in various scenarios. For example, when there are many highly correlated features, or more features than data points, L2-regularization is often preferable. In order to address the shortcomings of L1-regularization, elastic-net regularization~\cite{zou2005regularization} uses both L1 and an L2-regularization:
$    \frac{1}{n}\sum_{i=1}^n \ell(A^S(\mathbf{x}_i), y_i) + \lambda(\gamma||\textbf{w}_S||_2^2+\eta||\textbf{w}_S||_1).
$

Usually, $\eta=1-\gamma$ for $\gamma\leq1$~\cite{zou2005regularization}. Elastic-net regularization allows for the retention of strong convexity, due to L2 regularization, while also encouraging sparsity.  Elastic-net regularized algorithms satisfy $\beta$-uniform stability with
$
    \beta \leq \frac{2L^2\kappa}{n\lambda\gamma}
$~\cite{xu2011sparse}.

Elastic-net has been applied to generalized linear regression~\cite{friedman2010regularization}, logistic regression and support vector machines~\cite{wang2006doubly} in scenarios where sparsity is a desired property of the resulting algorithm, for example in medical applications~\cite{yu2014differentially}.

\begin{algorithm}
\caption{Private Elastic-Net via stability}\label{alg:elastic}
\begin{algorithmic}[1]
  \scriptsize
  \STATE Input: $S=\{(\mathbf{x}_i, y_i)\}$ for $i\in[1:n]$, loss function $\ell$, uniform stability $\beta= \frac{2L^2\kappa}{n\lambda\gamma}$ and privacy level $\epsilon$.
  \STATE  $\mathbf{w}_S \leftarrow $ minimizer of $\frac{1}{n}\sum_{i=1}^n \ell(A^S(\mathbf{x}_i), y_i) + \lambda(\gamma||\textbf{w}_S||_2^2+\eta||\textbf{w}_S||_1)$
  \STATE Return\footnotemark: $\mathbf{w}_S+\nu$ where $\nu\sim Lap\left(\frac{ \sqrt{(2\beta)/(\lambda\gamma)}}{\epsilon}\right)$ .
\end{algorithmic}
\end{algorithm}

\footnotetext{ Algorithm~\ref{alg:elastic} uses $L_2$-sensitivity due to~\cite{dwork2006calibrating, wu2017bolt}, this can be changed to L1-sensitivity with a factor of $\sqrt{d}$. }
\begin{corollary}\label{cor:elastic}
For any elastic-net regularized algorithm with convex loss $\ell$, Algorithm~\ref{alg:elastic} satisfies $\epsilon$-differential privacy.
\end{corollary}

\myparagraph{Private Feature Selection with Elastic-Net}\label{private-feature-selection}
Using Algorithm~\ref{alg:elastic}, we can obtain private regression, classification and feature selection algorithms using the uniform stability properties of elastic-net regularized loss functions.

In non-private elastic nets, feature decisions $f_i$ are made based on their weights $w_i$ as: $ f_i= 0$ if $w_i = 0$, and $f_i = 1$ otherwise.
That is, $f_i$ is selected for use iff it has a non-zero weight~\cite{zou2005regularization}.

In the private version, zero weights may be perturbed by noise. Private feature selection can be performed by obtaining differentially private model weights, as in Corollary~\ref{cor:elastic}, and then setting those weights with absolute value below a specified threshold $T$ to 0.

The following Lemma bounds the probability of a feature decision differing between private and non-private versions.

\begin{Lemma}\label{lemma:elastic}
Suppose we have a threshold $T > 0$, weight $w_i$ and non-private and private feature decisions $f_i$ and $f^{priv}_i$ for all $i\in [1, d]$. Then,
\mbox{$
    P[f_i\neq f^{priv}_i] = \exp\left(-\frac{\epsilon \abs{T - \abs{w_i} } \lambda \eta \sqrt{n}}{2L\sqrt{\kappa}}\right).
$}
As a direct consequence, for $\sqrt{n}\geq L\sqrt{\kappa}(\epsilon\lambda\eta)^{-1}$, $P[f_i\neq f^{priv}_i]\leq e^{-2(\abs{T - \abs{w_i}})}$.
\end{Lemma}


\section{Experimental Results\vspace*{-2mm}}\label{sec:exps}
Our empirical evaluation focuses on the performance of private elastic net regularized classifiers (see Algorithm~\ref{alg:elastic}) and the influence of stability on privacy noise and accuracy. We also show the effect of private elastic net on feature selection.

\myparagraph{Experimental set up.} The experiments used the Scikit-learn implementation \cite{pedregosa2011scikit} of the elastic net regularized logistic regression, which operates via SGD. Specifically, the weight of the $L_1$ and $L_2$ penalty were $0.15$ and $0.85$. We used $10$ seeded train-test splits with an $80/20\%$ split ratio to show mean performance metrics on the test set with standard deviation. As a pre-processing step we reduced the dimensionality of the features with Truncated SVD to generate 32 dimensional feature matrices, which were standardized column-wise.
\pgfplotsset{every axis title/.append style={at={(0.5,-0.55)}}}
\begin{figure}[h!]
\centering\vspace*{-2mm}
\input{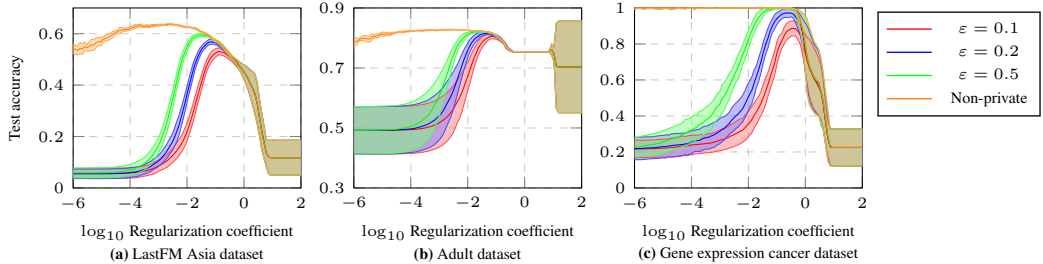}
\caption{Test accuracy of the private and non-private models on the \textit{LastFM Asia}, \textit{Adults}, and \textit{Gene expression cancer} datasets. The privacy noise is added optimally as a function of regularization constant according to Algorithm \ref{alg:elastic}. Accuracy increases with increasing stability. }\label{fig:regularization}
\vspace*{-3mm}
\end{figure}

\myparagraph{Datasets.} We utilized publicly available datasets with binary and multiclass classification tasks. The \textit{LastFM Asia dataset} \cite{snapnets, rozemberczki2020characteristic} contains users of the streaming service and the musicians these people liked. The related task is to predict the country of origin for the streamers. The \textit{Adult dataset} \cite{adult_dataset} was extracted from a census database. The classifier has to forecast the income category (low and high) of individuals using socio-economic indicators.  The \textit{Gene expression cancer dataset}  \cite{gene_cancer_dataset} contains cancerous tissue samples. Using gene expression measurements in the tissue, the task is to predict the type of cancer.

\myparagraph{Evaluation of Stability-Optimized Noise Tuning.} Figure~\ref{fig:regularization} shows the test accuracy of private and non-private models as a function of the regularization coefficient. For each privacy level and regularization coefficient pair, the privacy noise was set optimally using Algorithm \ref{alg:elastic}. The increase of the regularization decreases the sensitivity of the private models, and thus requires less noise to achieve the same privacy guarantee. Excessive weight regularization naturally degrades classification accuracy. The results confirm the idea from Section~\ref{sec:tradeoff} that using privacy enhanced by stability can be of net benefit in reducing error, while retaining same level of privacy. 

\begin{figure}[h!]
\centering\vspace*{-2mm}
\hspace*{-4mm}\input{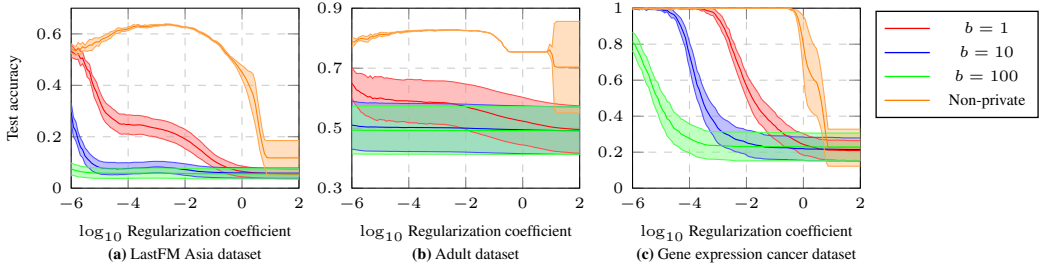}
\caption{Test accuracy of the private and non-private models on the \textit{LastFM Asia}, \textit{Adults}, and \textit{Gene expression cancer} datasets. The privacy noise is added with Laplace distribution of constant scale $b$ which is found to be substantially inferior to tuning the noise to stability.}\label{fig:regularization_suboptimal}\vspace*{-4mm}
\end{figure}
Figure~\ref{fig:regularization_suboptimal} shows the effect of not using the relation between privacy and stability and using a fixed noise level. For example, without knowledge of the implications of stability on sensitivity, one may choose to a fixed noise level given by a fixed scale $b$ to of the Laplace distribution. The result is that accuracy in decreases quickly with increasing regularization.


\myparagraph{Private Feature Selection.} Elastic-net regularization has a natural tendency to generate sparse model weights, and is thus useful for feature selection. We use here the idea described in Section~\ref{private-feature-selection} but set the threshold dynamically based on the standard deviation of the privacy noise parameter.

\begin{figure}[ht!]
\centering\vspace*{-2mm}
\hspace*{-4mm}\input{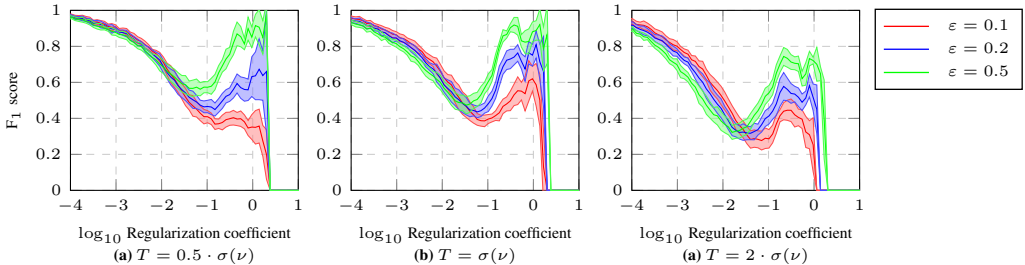}
\caption{The effect of artificial private model sparsification on the classifier weight structure similarity for the \textit{Gene expression} dataset.  
}\label{fig:dynamic_weight_modulation}\vspace*{-3mm}
\end{figure}

We compared the features selected by private and non-private models and the corresponding $F1$ scores for the gene expression dataset are shown in Figure \ref{fig:dynamic_weight_modulation}. At each value of regularization $\lambda$, the threshold $T$ was set based on the corresponding scale of noise. At extremely low $\lambda$, low weights are rare. As the elastic net takes effect, the selected features start to differ. After a value of $0.1$, the lower noise scale has noticeable effect and causes the selected feature vector similarity to rise. Figure~\ref{fig:weight_cutoff_1} in Appendix~\ref{appendix-experiments} shows results for static cutoffs. \vspace*{-2mm}



\section{Conclusion}

The results provided in this work directly link stability amplification to privacy amplification, further motivating the study of the uniform stability properties. This motivation applies even to those models which did not require stability results to guarantee generalization, which is the usual context for the study of uniformly stable algorithms. These results also suggest further study of actions which do not explicitly add random noise, but instead amplify the stability of the underlying learning algorithm as a means of improving the privacy-utility trade-off. Sampling and shuffling, are some examples that have been studied in similar contexts~\cite{balle2018privacy, erlingsson2019amplification}. 
Other directions for future work include weak convex and non-convex losses.

\section*{Broader Impact}

This work belongs to the general area of private machine learning~\cite{abadi2016deep,chaudhuri2011differentially, ji2014differential}. The objective being to perform machine learning in a way that the final output does not reveal too much about the data actually used in the input. Privacy is a major social concern in modern computing and machine learning~\cite{fernandez2013security, fiander2016house,  vellido2019societal}, and research in private learning algorithms are meant to mitigate those concerns to some extent. Other than the obvious societal benefits, developments in private learning can help development of better trained models, since training can be performed on sensitive data. This protection can also reassure more people to contribute their data to develop learning based systems. 

In this paper, we have shown that for algorithms with known stability bounds, privacy can be achieved at a smaller loss of accuracy. Thus, equally private algorithms will have better utility under these circumstances. This effect is particularly useful in applications with small training sets, which is often the case for privacy sensitive domains. Medical data is one such area where datasets are often small, and privacy is critical~\cite{haas2011aspects,vellido2019societal}. Since many modifications to learning algorithms, such as dropout~\cite{srivastava14a}, averaging and batch training that are commonly applied to training are known to be stable~\cite{hardt2015}, this result implies that such private versions of the algorithms (both current and future ones) can now claim better privacy. Though care should be taken that the results in some such works imply average stability~\cite{hardt2015, jordon2019differentially, verma2019stability, mou2018sgld, charles2018stability}, while our analysis requires a worst-case uniform stability bound~\cite{wu2017bolt, liu2016algorithm, xu2011sparse}. 

Our result can be applied in the form of ``output perturbation'' to computed models~\cite{dwork2006calibrating}, which means that it is easy for the non-expert to compute models using off-the-shelf libraries (such as Scikit-learn~\cite{pedregosa2011scikit}, which we used in experiments), and then impose differential privacy on the models. We hope this will help boost the popularity of private machine learning. 

A potential risk in the use of private machine learning is that the implications of a probabilistic guarantee like differential privacy is not the most intuitive to the lay person, and as a result, there is the risk of a gap between a citizen’s expectation of privacy and what an analysis like ours guarantees. The label of a private algorithm may give a false impression of \emph{absolute} protection, where the guarantee is really probabilistic, and the absolute level of privacy may depend on factors and adversarial knowledge outside the model.  

The limitations of the work presented here include the use of empirical results from a relatively small number of tasks and the reporting of overall classification accuracy as a performance metric, which can obscure differences in performance between classes. The experimental results satisfy $\epsilon$-differential privacy for each training run, however hyperparameter tuning was not performed privately~\cite{chaudhuri2011differentially}, which should be the case in deployed differentially private systems. Private hyperparameter tuning was not used as our experiments were intended to show the relationship between regularization and private model performance with the non-private results also presented for comparison. However, if used in practice, private hyperparameeter tuning is necessary.

This paper was intended to establish the theoretical concept. The practical adoption and evaluation will require further work.

\bibliographystyle{acm}
\bibliography{privacy}

\begin{thebibliography}{10}

\bibitem{abadi2016deep}
{\sc Abadi, M., Chu, A., Goodfellow, I., McMahan, H.~B., Mironov, I., Talwar,
  K., and Zhang, L.}
\newblock Deep learning with differential privacy.
\newblock In {\em Proceedings of the 2016 ACM SIGSAC Conference on Computer and
  Communications Security\/} (2016), pp.~308--318.

\bibitem{balle2018privacy}
{\sc Balle, B., Barthe, G., and Gaboardi, M.}
\newblock Privacy amplification by subsampling: Tight analyses via couplings
  and divergences.
\newblock In {\em Advances in Neural Information Processing Systems\/} (2018),
  pp.~6277--6287.

\bibitem{bassily2019private}
{\sc Bassily, R., Feldman, V., Talwar, K., and Thakurta, A.~G.}
\newblock Private stochastic convex optimization with optimal rates.
\newblock In {\em Advances in Neural Information Processing Systems\/} (2019),
  pp.~11279--11288.

\bibitem{bassily2014private}
{\sc Bassily, R., Smith, A., and Thakurta, A.}
\newblock Private empirical risk minimization, revisited.
\newblock {\em rem 3\/} (2014), 19.

\bibitem{bousquet2001}
{\sc Bousquet, O., and Elisseeff, A.}
\newblock Algorithmic stability and generalization performance.
\newblock In {\em Advances in Neural Information Processing Systems 13\/}
  (Cambridge, MA, USA, Apr. 2001), Max-Planck-Gesellschaft, MIT Press,
  pp.~196--202.

\bibitem{bousquet2002stability}
{\sc Bousquet, O., and Elisseeff, A.}
\newblock Stability and generalization.
\newblock {\em Journal of machine learning research 2}, Mar (2002), 499--526.

\bibitem{charles2018stability}
{\sc Charles, Z., and Papailiopoulos, D.}
\newblock Stability and generalization of learning algorithms that converge to
  global optima.
\newblock In {\em Proceedings of the International Conference on Machine
  Learning\/} (2018).

\bibitem{chaudhari2019entropy}
{\sc Chaudhari, P., Choromanska, A., Soatto, S., LeCun, Y., Baldassi, C.,
  Borgs, C., Chayes, J., Sagun, L., and Zecchina, R.}
\newblock Entropy-sgd: Biasing gradient descent into wide valleys.
\newblock {\em Journal of Statistical Mechanics: Theory and Experiment 2019},
  12 (2019), 124018.

\bibitem{chaudhuri2009privacy}
{\sc Chaudhuri, K., and Monteleoni, C.}
\newblock Privacy-preserving logistic regression.
\newblock In {\em Advances in neural information processing systems\/} (2009),
  pp.~289--296.

\bibitem{chaudhuri2011differentially}
{\sc Chaudhuri, K., Monteleoni, C., and Sarwate, A.~D.}
\newblock Differentially private empirical risk minimization.
\newblock {\em Journal of Machine Learning Research 12}, Mar (2011),
  1069--1109.

\bibitem{chen2018stability}
{\sc Chen, Y., Jin, C., and Yu, B.}
\newblock Stability and convergence trade-off of iterative optimization
  algorithms.
\newblock {\em arXiv preprint arXiv:1804.01619\/} (2018).

\bibitem{dozat2016incorporating}
{\sc Dozat, T.}
\newblock Incorporating nesterov momentum into adam.

\bibitem{Dwork06differentialprivacy}
{\sc Dwork, C.}
\newblock Differential privacy.
\newblock ICALP, pp.~1--12.

\bibitem{dwork2006calibrating}
{\sc Dwork, C., McSherry, F., Nissim, K., and Smith, A.}
\newblock Calibrating noise to sensitivity in private data analysis.
\newblock In {\em Theory of cryptography conference\/} (2006), Springer,
  pp.~265--284.

\bibitem{elisseeff2005stability}
{\sc Elisseeff, A., Evgeniou, T., and Pontil, M.}
\newblock Stability of randomized learning algorithms.
\newblock {\em Journal of Machine Learning Research 6}, Jan (2005), 55--79.

\bibitem{erlingsson2019amplification}
{\sc Erlingsson, {\'U}., Feldman, V., Mironov, I., Raghunathan, A., Talwar, K.,
  and Thakurta, A.}
\newblock Amplification by shuffling: From local to central differential
  privacy via anonymity.
\newblock In {\em Proceedings of the Thirtieth Annual ACM-SIAM Symposium on
  Discrete Algorithms\/} (2019), SIAM, pp.~2468--2479.

\bibitem{feldman2018generalization}
{\sc Feldman, V., and Vondrak, J.}
\newblock Generalization bounds for uniformly stable algorithms.
\newblock In {\em Advances in Neural Information Processing Systems\/} (2018),
  pp.~9747--9757.

\bibitem{fernandez2013security}
{\sc Fern{\'a}ndez-Alem{\'a}n, J.~L., Se{\~n}or, I.~C., Lozoya, P. {\'A}.~O.,
  and Toval, A.}
\newblock Security and privacy in electronic health records: A systematic
  literature review.
\newblock {\em Journal of biomedical informatics 46}, 3 (2013), 541--562.

\bibitem{fiander2016house}
{\sc Fiander, S., and Blackwood, N.}
\newblock House of commons science and technology committee: Robotics and
  artificial intelligence: Fifth report of session 2016--17.

\bibitem{friedman2010regularization}
{\sc Friedman, J., Hastie, T., and Tibshirani, R.}
\newblock Regularization paths for generalized linear models via coordinate
  descent.
\newblock {\em Journal of statistical software 33}, 1 (2010), 1.

\bibitem{gao2013stability}
{\sc Gao, W., and Xu, T.}
\newblock Stability analysis of learning algorithms for ontology similarity
  computation.
\newblock In {\em Abstract and Applied Analysis\/} (2013), vol.~2013, Hindawi.

\bibitem{gupta2016differentially}
{\sc Gupta, S.~K., Rana, S., and Venkatesh, S.}
\newblock Differentially private multi-task learning.
\newblock In {\em Pacific-Asia Workshop on Intelligence and Security
  Informatics\/} (2016), Springer, pp.~101--113.

\bibitem{haas2011aspects}
{\sc Haas, S., Wohlgemuth, S., Echizen, I., Sonehara, N., and M{\"u}ller, G.}
\newblock Aspects of privacy for electronic health records.
\newblock {\em International journal of medical informatics 80}, 2 (2011),
  e26--e31.

\bibitem{hardt2015}
{\sc Hardt, M., Recht, B., and Singer, Y.}
\newblock Train faster, generalize better: Stability of stochastic gradient
  descent.
\newblock {\em CoRR abs/1509.01240\/} (2015).

\bibitem{jain2015drop}
{\sc Jain, P., Kulkarni, V., Thakurta, A., and Williams, O.}
\newblock To drop or not to drop: Robustness, consistency and differential
  privacy properties of dropout.
\newblock {\em arXiv preprint arXiv:1503.02031\/} (2015).

\bibitem{jain2014near}
{\sc Jain, P., and Thakurta, A.~G.}
\newblock (near) dimension independent risk bounds for differentially private
  learning.
\newblock In {\em International Conference on Machine Learning\/} (2014),
  pp.~476--484.

\bibitem{ji2014differential}
{\sc Ji, Z., Lipton, Z.~C., and Elkan, C.}
\newblock Differential privacy and machine learning: a survey and review.

\bibitem{jordon2019differentially}
{\sc Jordon, J., Yoon, J., and van~der Schaar, M.}
\newblock Differentially private bagging: Improved utility and cheaper privacy
  than subsample-and-aggregate.
\newblock In {\em Advances in Neural Information Processing Systems\/} (2019),
  pp.~4325--4334.

\bibitem{kifer2012private}
{\sc Kifer, D., Smith, A., and Thakurta, A.}
\newblock Private convex empirical risk minimization and high-dimensional
  regression.
\newblock In {\em Conference on Learning Theory\/} (2012), pp.~25--1.

\bibitem{adult_dataset}
{\sc Kohavi, R.}
\newblock Scaling up the accuracy of naive-bayes classifiers: A decision-tree
  hybrid.
\newblock In {\em Proceedings of the Second International Conference on
  Knowledge Discovery and Data Mining\/} (1996), p.~202–207.

\bibitem{snapnets}
{\sc Leskovec, J., and Krevl, A.}
\newblock {SNAP Datasets}: {Stanford} large network dataset collection.
\newblock \url{http://snap.stanford.edu/data}, June 2014.

\bibitem{liu2017algorithmic}
{\sc Liu, T., Lugosi, G., Neu, G., and Tao, D.}
\newblock Algorithmic stability and hypothesis complexity.
\newblock In {\em Proceedings of the 34th International Conference on Machine
  Learning-Volume 70\/} (2017), JMLR. org, pp.~2159--2167.

\bibitem{liu2016algorithm}
{\sc Liu, T., Tao, D., Song, M., and Maybank, S.~J.}
\newblock Algorithm-dependent generalization bounds for multi-task learning.
\newblock {\em IEEE transactions on pattern analysis and machine intelligence
  39}, 2 (2016), 227--241.

\bibitem{mou2018sgld}
{\sc Mou, W., Wang, L., Zhai, X., and Zheng, K.}
\newblock Generalization bounds of sgld for non-convex learning: Two
  theoretical viewpoints.
\newblock In {\em Proceedings of the 31st Conference On Learning Theory\/}
  (06--09 Jul 2018), S.~Bubeck, V.~Perchet, and P.~Rigollet, Eds., vol.~75 of
  {\em Proceedings of Machine Learning Research}, PMLR, pp.~605--638.

\bibitem{mukherjee2006learning}
{\sc Mukherjee, S., Niyogi, P., Poggio, T., and Rifkin, R.}
\newblock Learning theory: stability is sufficient for generalization and
  necessary and sufficient for consistency of empirical risk minimization.
\newblock {\em Advances in Computational Mathematics 25}, 1-3 (2006), 161--193.

\bibitem{pedregosa2011scikit}
{\sc Pedregosa, F., Varoquaux, G., Gramfort, A., Michel, V., Thirion, B.,
  Grisel, O., Blondel, M., Prettenhofer, P., Weiss, R., Dubourg, V., et~al.}
\newblock Scikit-learn: Machine learning in python.
\newblock {\em the Journal of machine Learning research 12\/} (2011),
  2825--2830.

\bibitem{poggio2009sufficient}
{\sc Poggio, T., Rosasco, L., and Wibisono, A.}
\newblock Sufficient conditions for uniform stability of regularization
  algorithms.

\bibitem{rogers1978finite}
{\sc Rogers, W.~H., and Wagner, T.~J.}
\newblock A finite sample distribution-free performance bound for local
  discrimination rules.
\newblock {\em The Annals of Statistics\/} (1978), 506--514.

\bibitem{rozemberczki2020characteristic}
{\sc Rozemberczki, B., and Sarkar, R.}
\newblock Characteristic functions on graphs: Birds of a feather, from
  statistical descriptors to parametric models, 2020.

\bibitem{rubinstein2012learning}
{\sc Rubinstein, B.~I., Bartlett, P.~L., Huang, L., and Taft, N.}
\newblock Learning in a large function space: Privacy-preserving mechanisms for
  svm learning.
\newblock {\em Journal of Privacy and Confidentiality 4}, 1 (2012), 65--100.

\bibitem{santurkar2018does}
{\sc Santurkar, S., Tsipras, D., Ilyas, A., and Madry, A.}
\newblock How does batch normalization help optimization?
\newblock In {\em Advances in Neural Information Processing Systems\/} (2018),
  pp.~2483--2493.

\bibitem{scholkopf2001learning}
{\sc Scholkopf, B., and Smola, A.~J.}
\newblock {\em Learning with kernels: support vector machines, regularization,
  optimization, and beyond}.
\newblock MIT press, 2001.

\bibitem{singh2016swapout}
{\sc Singh, S., Hoiem, D., and Forsyth, D.}
\newblock Swapout: Learning an ensemble of deep architectures.
\newblock In {\em Advances in neural information processing systems\/} (2016),
  pp.~28--36.

\bibitem{song2017machine}
{\sc Song, C., Ristenpart, T., and Shmatikov, V.}
\newblock Machine learning models that remember too much.
\newblock In {\em Proceedings of the 2017 ACM SIGSAC Conference on Computer and
  Communications Security\/} (2017), pp.~587--601.

\bibitem{srivastava14a}
{\sc Srivastava, N., Hinton, G., Krizhevsky, A., Sutskever, I., and
  Salakhutdinov, R.}
\newblock Dropout: A simple way to prevent neural networks from overfitting.
\newblock {\em Journal of Machine Learning Research 15}, 56 (2014), 1929--1958.

\bibitem{tibshirani1996regression}
{\sc Tibshirani, R.}
\newblock Regression shrinkage and selection via the lasso.
\newblock {\em Journal of the Royal Statistical Society: Series B
  (Methodological) 58}, 1 (1996), 267--288.

\bibitem{vellido2019societal}
{\sc Vellido, A.}
\newblock Societal issues concerning the application of artificial intelligence
  in medicine.
\newblock {\em Kidney Diseases 5}, 1 (2019), 11--17.

\bibitem{verma2019stability}
{\sc Verma, S., and Zhang, Z.-L.}
\newblock Stability and generalization of graph convolutional neural networks.
\newblock In {\em Proceedings of the 25th ACM SIGKDD International Conference
  on Knowledge Discovery \& Data Mining\/} (2019), pp.~1539--1548.

\bibitem{wang2017}
{\sc Wang, D., Ye, M., and Xu, J.}
\newblock Differentially private empirical risk minimization revisited: Faster
  and more general.
\newblock In {\em Advances in Neural Information Processing Systems 30},
  I.~Guyon, U.~V. Luxburg, S.~Bengio, H.~Wallach, R.~Fergus, S.~Vishwanathan,
  and R.~Garnett, Eds. Curran Associates, Inc., 2017, pp.~2722--2731.

\bibitem{wang2017differentially}
{\sc Wang, D., Ye, M., and Xu, J.}
\newblock Differentially private empirical risk minimization revisited: Faster
  and more general.
\newblock In {\em Advances in Neural Information Processing Systems\/} (2017),
  pp.~2722--2731.

\bibitem{wang2006doubly}
{\sc Wang, L., Zhu, J., and Zou, H.}
\newblock The doubly regularized support vector machine.
\newblock {\em Statistica Sinica\/} (2006), 589--615.

\bibitem{wang2015privacy}
{\sc Wang, Y.-X., Fienberg, S., and Smola, A.}
\newblock Privacy for free: Posterior sampling and stochastic gradient monte
  carlo.
\newblock In {\em International Conference on Machine Learning\/} (2015),
  pp.~2493--2502.

\bibitem{Wangprivacy}
{\sc Wang, Y.-X., Lei, J., and Fienberg, S.~E.}
\newblock Learning with differential privacy: Stability, learnability and the
  sufficiency and necessity of erm principle.
\newblock {\em J. Mach. Learn. Res. 17}, 1 (Jan. 2016), 6353--6392.

\bibitem{gene_cancer_dataset}
{\sc Weinstein, J.~N., Collisson, E.~A., Mills, G.~B., Shaw, K. R.~M.,
  Ozenberger, B.~A., Ellrott, K., Shmulevich, I., Sander, C., Stuart, J.~M.,
  Network, C. G. A.~R., et~al.}
\newblock The cancer genome atlas pan-cancer analysis project.
\newblock {\em Nature genetics 45}, 10 (2013), 1113.

\bibitem{wu2017bolt}
{\sc Wu, X., Li, F., Kumar, A., Chaudhuri, K., Jha, S., and Naughton, J.}
\newblock Bolt-on differential privacy for scalable stochastic gradient
  descent-based analytics.
\newblock In {\em Proceedings of the 2017 ACM International Conference on
  Management of Data\/} (2017), pp.~1307--1322.

\bibitem{xu2011sparse}
{\sc Xu, H., Caramanis, C., and Mannor, S.}
\newblock Sparse algorithms are not stable: A no-free-lunch theorem.
\newblock {\em IEEE transactions on pattern analysis and machine intelligence
  34}, 1 (2011), 187--193.

\bibitem{yu2014differentially}
{\sc Yu, F., Rybar, M., Uhler, C., and Fienberg, S.~E.}
\newblock Differentially-private logistic regression for detecting multiple-snp
  association in gwas databases.
\newblock In {\em International Conference on Privacy in Statistical
  Databases\/} (2014), Springer, pp.~170--184.

\bibitem{zhang2020privacy}
{\sc Zhang, C., Hu, X., Xie, Y., Gong, M., and Yu, B.}
\newblock A privacy-preserving multi-task learning framework for face
  detection, landmark localization, pose estimation, and gender recognition.
\newblock {\em Frontiers in Neurorobotics 13\/} (2020), 112.

\bibitem{zou2005regularization}
{\sc Zou, H., and Hastie, T.}
\newblock Regularization and variable selection via the elastic net.
\newblock {\em Journal of the royal statistical society: series B (statistical
  methodology) 67}, 2 (2005), 301--320.

\end{thebibliography}


\appendix
\clearpage
\section{Proofs}

\begin{proof} \textbf{(Theorem~\ref{thm:ruben}) }

The proof is analogous to that of~\cite{rubinstein2012learning, bousquet2002stability}, repeated here for completeness.


As $A_S$ is defined to minimize $L(A, S)$, it follows that the partial derivative of $L(A, S)$ evaluated at $S$ will have value 0. This is implied by the necessary conditions from Karush-Kuhn-Tucker (KKT) multipliers:
\begin{equation}
\partial_A L^R(\as, S)=\partial_A L(\as, S) + \lambda||\mathbf{w}_S||=0
\end{equation}
\begin{equation}
\partial_A L^{R}(\asp, S)=\partial_A L(\asp, S) + \lambda||\mathbf{w}_{S'}||=0
\end{equation}

Construct an \textit{auxiliary risk function} as follows:
\begin{equation}
    \overline{L}(A)=  \langle\partial_A L(A^S, S)-\partial_A L(\asp, S') , \mathbf{w}_A - \mathbf{w}_{S'} \rangle + \lambda ||\mathbf{w}_A-\mathbf{w}_{S'}||
\end{equation}
Where, $\mathbf{w}_A$ represents the weights associated with some algorithm $A$ and $\mathbf{w}_{S'}$ represents the weights associated with the algorithm $A$ trained to minimze $L^R$ over the dataset $S'$.

Note that:
\begin{enumerate}

    \item The auxiliary risk function $ \overline{L}(A)$ is convex
    as the first term is linear, and the second quadratic.

   \item By construction, $ \overline{L}(\asp)=0$

   \item $ \overline{L}$ is minimzed by $A^S$, as the partial derivative is given by: \\
\begin{equation}
\begin{split}
\partial_A \overline{L} &= \partial_A L(A^S, S) - \partial_A L(\asp, S') + \lambda \mathbf{w}_A - \lambda \mathbf{w}_{S'} \\
      &= \partial_A L(A^S, S) + \lambda \mathbf{w}_A
\end{split}
\end{equation}
\end{enumerate}

Therefore, $ \overline{L}(\as) \leq 0$ as by the convexity of $ \overline{L}$ its inflection point must have a value less than or equal to the value of $ \overline{L}(\asp)$.


The first term of $ \overline{L}(A)$ can be simplified as follows, for $(\mathbf{x}_i, y_i)\in S$:

\begin{align*}
    n \overline{L}(A^S) &=n \langle \partial_A L(A^S, S)-\partial_A L(\asp, S') , \mathbf{w}_S - \mathbf{w}_{S'} \rangle\\
    & =\sum_{i=1}^{n}  \langle\partial_A l( A^S(\mathbf{x}_i),y_i)-\partial_A l( A^{S'}(\mathbf{x}_i'), y_i'),\mathbf{w}_S- \mathbf{w}_{S'} \rangle\\
    & =\sum_{i=1}^{n} \left(\left(l'( A^S(\mathbf{x}_i), y_i)-l'( A^{S'}(\mathbf{x}_i'), y_i')\right)\left(A_S(\mathbf{x}_i)-\asp(\mathbf{x}_i)\right)\right)\\
    &+ l'\left(  A^S(\mathbf{x}_n),y_n \right)\left(A_S(\mathbf{x}_n)-\asp(\mathbf{x}_n)\right)-l'\left(  \asp(\mathbf{x}_n'),y_n' \right)\left(A^S(\mathbf{x}_n')-\asp(\mathbf{x}_n')\right)\\
    & \geq l'\left( A^S(\mathbf{x}_n),y_n \right)\left(A_S(\mathbf{x}_n)-\asp(\mathbf{x}_n)\right)-l'\left(  \asp(\mathbf{x}_n'), y_n' \right)\left(A^S(\mathbf{x}_n')-\asp(\mathbf{x}_n')\right)\\
\end{align*}


Therefore, by combining with $ \overline{L}(\as) \leq 0$ we obtain:
\begin{equation}
    0 \geq n \overline{L}(A^S)\geq l'\left(  A^S(\mathbf{x}_n),y_n \right)\left(A_S(\mathbf{x}_n)-\asp(\mathbf{x}_n)\right)-l'\left(  \asp(\mathbf{x}_n'),y_n' \right)\left(A^S(\mathbf{x}_n')-\asp(\mathbf{x}_n')\right)
\end{equation}
Rearranging:
\begin{equation}
    -n \overline{L}(A^S)\geq l'\left(  A^S(\mathbf{x}_n),y_n \right)\left(A_S(\mathbf{x}_n)-\asp(\mathbf{x}_n)\right)-l'\left(  \asp(\mathbf{x}_n'),y_n' \right)\left(A^S(\mathbf{x}_n')-\asp(\mathbf{x}_n')\right)
\end{equation}
We obtain:
\begin{equation}
    n \overline{L}(A^S)\leq l'\left(  \asp(\mathbf{x}_n'),y_n' \right)\left(A^S(\mathbf{x}_n')-\asp(\mathbf{x}_n')\right)-l'\left(  A^S(\mathbf{x}_n),y_n \right)\left(A_S(\mathbf{x}_n)-\asp(\mathbf{x}_n)\right)
\end{equation}
As, $\frac{\lambda}{2}||\mathbf{w}_S-\mathbf{w}_{S'}||^2 \leq \langle\partial_A L(A^S, S)-\partial_A L(\asp, S') , \mathbf{w}_S - \mathbf{w}_{S'} \rangle + \lambda ||\mathbf{w}_S-\mathbf{w}_{S'}||^2$, then:
\begin{equation}
    \frac{n\lambda}{2}||\mathbf{w}_S-\mathbf{w}_{S'}||^2
\end{equation}
\begin{equation}
\leq l'\left( \asp(\mathbf{x}_n'), y_n' \right)\left(A^S(\mathbf{x}_n')-\asp(\mathbf{x}_n')\right)-l'\left(  A^S(\mathbf{x}_n) ,y_n\right)\left(A_S(\mathbf{x}_n)-\asp(\mathbf{x}_n)\right)
\end{equation}


By the Lipschitz continuity of $l$ this results in:
\begin{equation}
 \frac{n\lambda}{2}||\mathbf{w}_S-\mathbf{w}_{S'}||^2\leq 2L||A_S-A_{S'}||_\infty
 \label{infnorm}
 \end{equation}

 Using the reproducing property alongside the Cauchy-Schwartz inequality, for each $\mathbf{x}$:

 \begin{align}
     |A_S(\mathbf{x})-A_{S'}(\mathbf{x})| &=|\langle \phi(\mathbf{x}), \mathbf{w}_S-\mathbf{w}_{S'}\rangle|\\
     &\leq ||\phi(\mathbf{x})||_2||\mathbf{w}_S-\mathbf{w}_{S'}||_2\\
       &= \sqrt(k(\mathbf{x}, \mathbf{x})||\mathbf{w}_S-\mathbf{w}_{S'}||_2\\
       &\leq \kappa ||\mathbf{w}_S-\mathbf{w}_{S'}||_2
 \end{align}

 Therefore, combining with (\ref{infnorm}):
 \begin{equation}
     ||\mathbf{w}_S-\mathbf{w}_{S'}||_2 \leq \frac{4L\kappa}{n\lambda}
 \end{equation}

To instead obtain the uniform stability, use the Lipschitz continuity of $\ell$ to obtain:
 \begin{equation}
     ||\ell( A^S(x_i),y)-\ell( A^{S'}(x_i), y)|| \leq L  ||\mathbf{w}_S-\mathbf{w}_{S'}||_k \leq \frac{4L^2\kappa^2}{n\lambda}
 \end{equation}
 \end{proof}
 \begin{proof}\textbf{(Theorem~\ref{thm:stability})}

By the strong convexity of $\ell$, for all $x\in S$:
\[
    \ell(A^S, x)-\ell(A^{S'}, x) \geq \nabla \ell(A^{S'}, x)^T(\textbf{w}_S-\textbf{w}_{S'})+\frac{\lambda}{2}||\textbf{w}_S-\textbf{w}_{S'}||_2^2
\]
Which implies, by the uniform stability of $\ell$:
\[
\beta \geq \nabla \ell(A^{S'}, x)^T(\textbf{w}_S-\textbf{w}_{S'})+\frac{\lambda}{2}||\textbf{w}_S-\textbf{w}_{S'}||_2^2
\]
By the convexity of $\ell$, we have that $\nabla \ell(A^{S'}, x)^T(\textbf{w}_S-\textbf{w}_{S'})\geq0$ for minimizer $\textbf{w}_{S'}$, therefore:
\[
\beta \geq \frac{\lambda}{2}||\textbf{w}_S-\textbf{w}_{S'}||_2^2
\]
\[
||\textbf{w}_S-\textbf{w}_{S'}||_2 \leq \sqrt{\frac{2\beta}{\lambda}}
.\]
\end{proof}

\begin{proof}\textbf{(Corollary~\ref{cor:enhance})}
 Due to Theorem 7, noise $r_1$ drawn from the following distribution is sufficient to provide $\epsilon$-differential privacy for algorithm $A_1$ which is $\beta_1$-uniformly stable:
 \begin{equation}
     r_1\sim Lap\left(\frac{\sqrt{2\beta_1}}{\epsilon\sqrt{\lambda}}\right)
 \end{equation}
 In comparison, for $A_2$, noise $r_2$ drawn as follows will suffice:
  \begin{equation}
     r_2\sim Lap\left(\frac{\sqrt{2\beta_2}}{\epsilon\sqrt{\lambda}}\right)
 \end{equation}
 Note that, for $\epsilon'=\sqrt{\frac{\beta_2}{\beta_1}}\epsilon$:
 \begin{equation}
   Lap\left(\frac{\sqrt{2\beta_1}}{\epsilon\sqrt{\lambda}}\right)=Lap\left(\frac{\sqrt{2\beta_2}}{\epsilon'\sqrt{\lambda}}\right)
 \end{equation}
 Therefore, algorithm $A_2$ with noise drawn according to $r_1$ satisfies $\sqrt{\frac{\beta_2}{\beta_1}}\epsilon$-differential privacy
\end{proof}
 \begin{proof}\textbf{(Remark~\ref{rem:conv})}
 \begin{align}
    ||\widetilde{\mathbf{w}}_S-\widetilde{\mathbf{w}}_{S'}||_2
    &\leq 2\delta_{conv}+||\mathbf{w}_{S}-\mathbf{w}_{S'}||_2
\end{align}

For $\lambda$-strongly convex functions, denote the loss achieved by $\widetilde{\mathbf{w}}_S$ as $ \widetilde{\ell}(A^S, s)$, then for all $s\in S$ and minimizer $\mathbf{w}_S$:
\begin{align}
    \widetilde{\ell}(A^S, s)-\ell(A^S, s) &\geq \nabla\ell(A^S, s)^T(\widetilde{\mathbf{w}}_S-\mathbf{w}_S)+||\widetilde{\mathbf{w}}_S-\mathbf{w}_S||_2^2 \\
    &\geq ||\widetilde{\mathbf{w}}_S-\mathbf{w}_S||_2^2
\end{align}
Therefore, if $|\widetilde{\ell}(A^S, s)-\ell(A^S, s)|\leq \gamma_{conv}$:
\begin{equation}
    ||\widetilde{\mathbf{w}}_S-\mathbf{w}_S||_2 \leq \sqrt{|\widetilde{\ell}(A^S, s)-\ell(A^S, s)|} \leq \sqrt{\gamma_{conv}}
\end{equation}
 \end{proof}

\begin{proof} \textbf{(Lemma~\ref{lemma:priverror})}
\begin{equation}
   \delta_{priv}= \mathbb{E}[|\frac{1}{n}\sum_{i=1}^n \ell(\mathbf{w}_{priv}, z_i)-\frac{1}{n}\sum_{i=1}^n \ell(\mathbf{w}, z_i)|]\leq \frac{1}{n}\sum_{i=1}^n\mathbb{E}[|\ell(\mathbf{w}_{priv}, z_i) -\ell(\mathbf{w}, z_i)|]
\end{equation}
By the monotonicity of expectation, and the lipschitz continuity of $\ell$ such that $|\ell(\mathbf{w}_{priv}, z_i) -\ell(\mathbf{w}, z_i)|\leq L|\mathbf{w}_{priv}- \mathbf{w}|$ we have that for all $z_i$:
\begin{equation}
    \frac{1}{n}\sum_{i=1}^n\mathbb{E}[|\ell(\mathbf{w}_{priv}, z_i) -\ell(\mathbf{w}, z_i)|] \leq \frac{L}{n}\sum_{i=1}^n\mathbb{E}[|\mathbf{w}_{priv} -\mathbf{w}|]
\end{equation}
As $\mathbf{w}_{priv}=\mathbf{w}+r$, where $r$ is a $d$-dimenstional vector of random noise s.t. $r_i\sim Lap(0, \frac{\Delta f}{\epsilon})$:
\begin{equation}
    \frac{L}{n}\sum_{i=1}^n\mathbb{E}[|\mathbf{w}_{priv} -\mathbf{w}|] =  \frac{L}{n}\sum_{i=1}^n\mathbb{E}[|r|] = L\mathbb{E}[|r|]
\end{equation}
By the triangle inequality and the monotonicity of expectation:
\begin{equation}
   L\mathbb{E}[|r|]=L\mathbb{E}[|r_i|+...+|r_d|]\leq Ld \mathbb{E}[|r_i|]
\end{equation}
As $r_i\sim Lap(0, \frac{\Delta f}{\epsilon})$, $|r_i|\sim Exponential( \frac{\epsilon }{\Delta f})$ and $E[|r_i|]=\frac{\Delta f}{\epsilon}.$
\begin{equation}
    \delta_{priv}\leq \frac{Ld}{\epsilon }\sqrt{\frac{2\beta}{\lambda}}
\end{equation}
\end{proof}

\begin{proof}\textbf{(Lemma~\ref{lemma:elastic})}
Denote the random noise added to weight $w_i$ by $r_i$.

\underline{Case 1: $f_i=0$.}
 In this scenario, the feature decision changes if the Laplace noise $r_i$ results in a private weight vector with absolute value larger than $T$. By a direct application of the cumulative distribution function of the Laplace distribution, with mean $0$ and scale $\sqrt{\frac{2\beta}{\lambda}}$ where $\beta= \frac{2L^2\kappa}{n\lambda\gamma}$, we obtain:
\[P[f_i\neq f^{priv}_i] = 1- P[|r_i|\leq T] = \exp\left(-\frac{\epsilon T \lambda \eta \sqrt{n}}{2L\sqrt{\kappa}}\right)\]

\underline{Case 2: $f_i \neq 0$.}
In this case, the feature decision changes if the Laplace noise is such that the private weight has absolute value $\leq T$.
\[P[f_i\neq f^{priv}_i] = 1-P[|r_i|\leq|T-|w_i||)]=\exp\left(-\frac{\epsilon |T-|w_i|| \lambda \eta \sqrt{n}}{2L\sqrt{\kappa}}\right)\]

Combining these cases, we obtain the result.
\end{proof}
\subsection{Average-Case Stability Results}\label{sec:averagestability}

\begin{table}[h]
\begin{tabular}{l||c|c|c|c}
Method & Parameters & Stability &  \begin{tabular}[c]{@{}l@{}} Prev. Work  \\ (Output Pert.) \end{tabular} & \begin{tabular}[c]{@{}l@{}} Prev. Work  \\ (Obj. Pert.) \end{tabular}\\ \hline \hline

    \hline

       $\star$ SGD - Dropout~\cite{hardt2015}     & Rate $s$ &   $O(s)$      & -&\cite{jain2015drop}            \\
       $\star$ SGD - Swapout~\cite{singh2016swapout}     & $L$ & $O(L)$      & -                 \\
$\star$ SGLD- Agg. Step Size~\cite{ mou2018sgld} & $T_k$ & $O(\sqrt{T_k})$& -& \cite{ wang2015privacy} \\
    $\star$ SGLD- Inverse Temp.~\cite{mou2018sgld} & $\beta$  & $O(\sqrt{\beta})$ &-&\cite{wang2015privacy} \\
    $\star$ RCD~\cite{charles2018stability} &$\lambda$ &$O(\frac{1}{\lambda})$&- & - \\
    $\star$ SVRG~\cite{charles2018stability} & $\gamma,T$ &$O((\frac{2L\gamma}{1-2L\gamma})^T)$&- &\cite{wang2017differentially}\\
     $\star$ Entropy-SGD~\cite{chaudhari2019entropy}    &$\alpha$&$O(\cdot)$ & - &- \\
    $\star$ 1-layer Graph-CNN~\cite{verma2019stability}& $\lambda_G^{max}$& $O(\lambda_G^{max})$ &- &-\\
    $\star$ Bagging~\cite{elisseeff2005stability}  &$m$& $O(\frac{1}{m})$ & \cite{jordon2019differentially} &-\\

\end{tabular}
\vspace{1mm}
\caption{Uniformly stable algorithms and the parameters which can be used to control their stability. Related work included in the last columns indicates if the relationship to the described parameter has been previously used in the context of output perturbation or objective perturbation. $\star$ indicates average-case uniform stability results, which require worst-case counterparts.}
\end{table}

\clearpage

\section{Additional Experiments}\label{appendix-experiments}
\subsection{Suboptimal Noise Mixing with Changing Scale}
\begin{figure}[h!]
\centering
\input{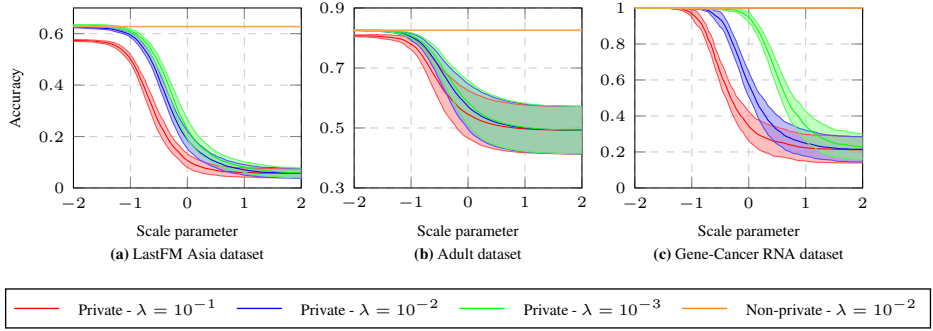}
\caption{Test accuracy of the private and non-private models on the \textit{LastFM Asia}, \textit{Adults}, and \textit{Gene expression cancer} datasets. The privacy noise is added suboptimally according to a zero mean, changing scale Laplace distribution with fixed regularization coefficients.}
\end{figure}
\subsection{Non-dynamic Weight Cutoff}
\begin{figure}[ht!]
\centering
\input{images/weight_modulation}
\caption{The effect of artificial private model sparsification on the classifier weight structure similarity for the \textit{Gene expression cancer} dataset. When the weight regularization is high, dropping small noisy weights ensures that the private and non-private models have similar sparsity structure.}\label{fig:weight_cutoff_1}
\end{figure}

\begin{figure}[ht!]
\centering
\input{images/weight_cutoff}
\caption{The effect of artificial private model sparsification on the classifier weight structure similarity for the \textit{Gene expression cancer} dataset. When the weight regularization is high, dropping small noisy weights ensures that the private and non-private models have similar sparsity structure.}\label{fig:weight_cutoff_2}
\end{figure}


\end{document}